\documentclass[letterpaper, 10 pt, conference]{ieeeconf}  

\IEEEoverridecommandlockouts                              

\usepackage{graphicx}
\usepackage{amsmath} 
\usepackage{amssymb}  
\usepackage[utf8]{inputenc}
\usepackage{balance}

\title{\LARGE \bf
Real Time Dense Depth Estimation by Fusing Stereo with Sparse Depth Measurements
}

\author{Shreyas S. Shivakumar, Kartik Mohta, Bernd Pfrommer, Vijay Kumar and Camillo J. Taylor
\thanks{We gratefully acknowledge the support of DARPA grants HR001151626 and HR0011516850, ARO grant W911NF-13-1-0350, ONR grant N00014-07-1-0829 and USDA grant 2015-67021-23857}
\thanks{Shreyas S. Shivakumar, Kartik Mohta, Bernd Pfrommer, Vijay Kumar and Camillo J. Taylor are with the GRASP Laboratory, School of Engineering and Applied Sciences,
        University of Pennsylvania, Philadelphia PA 19104
        {\tt\small \{sshreyas,kmohta,bpfrommer,kumar,cjtaylor\}
        @seas.upenn.edu}}%
}

\begin{document}

\maketitle
\thispagestyle{empty}
\pagestyle{empty}

\begin{abstract}

We present an approach to depth estimation that fuses information from a stereo pair with sparse range measurements derived from a LIDAR sensor or a range camera.
The goal of this work is to exploit the complementary strengths of the two sensor modalities, the accurate but sparse range measurements and the ambiguous but dense stereo information. These two sources are effectively and efficiently fused by combining ideas from anisotropic diffusion and semi-global matching.

We evaluate our approach on the KITTI 2015 and Middlebury 2014 datasets, using randomly sampled ground truth range measurements as our sparse depth input. We achieve significant performance improvements with a small fraction of range measurements on both datasets. We also provide qualitative results from our platform using the PMDTec Monstar sensor. Our entire pipeline runs on an NVIDIA TX-2 platform at 5Hz on 1280$\times$1024 stereo images with 128 disparity levels.
\end{abstract}

\section{INTRODUCTION}

Accurate real-time dense depth estimation is a challenging task for mobile robots.  Most often, a combination of sensors is used to improve performance. Sensor fusion is the broad category of combining various on-board sensors to produce better measurement estimates. These sensors are combined to compliment each other and  overcome individual shortcomings. We focus on the fusion of high resolution image data with low resolution depth measurements, which is a common method of obtaining dense 3D information. 

Passive stereo cameras are a popular choice for 3D perception in mobile robots, able to generate dense depth estimates that are readily scaled by increasing the resolution of the sensors used. However, stereo depth estimation algorithms are typically dependent upon visual cues and scene texture and can struggle to assign disparities in regions that contain of uniform patches, blurred regions and large illumination changes. Depending on the resolution and performance desired, dense stereo based depth estimation can be computationally demanding on compute constrained robot platforms. However, embedded hardware accelerators such as the Nvidia TX-2 can exploit the parallelism inherent in the stereo matching algorithms making these approaches practical for robotic applications \cite{hernandez2016embedded}.

\begin{figure}[t!]
\includegraphics[height=80px]{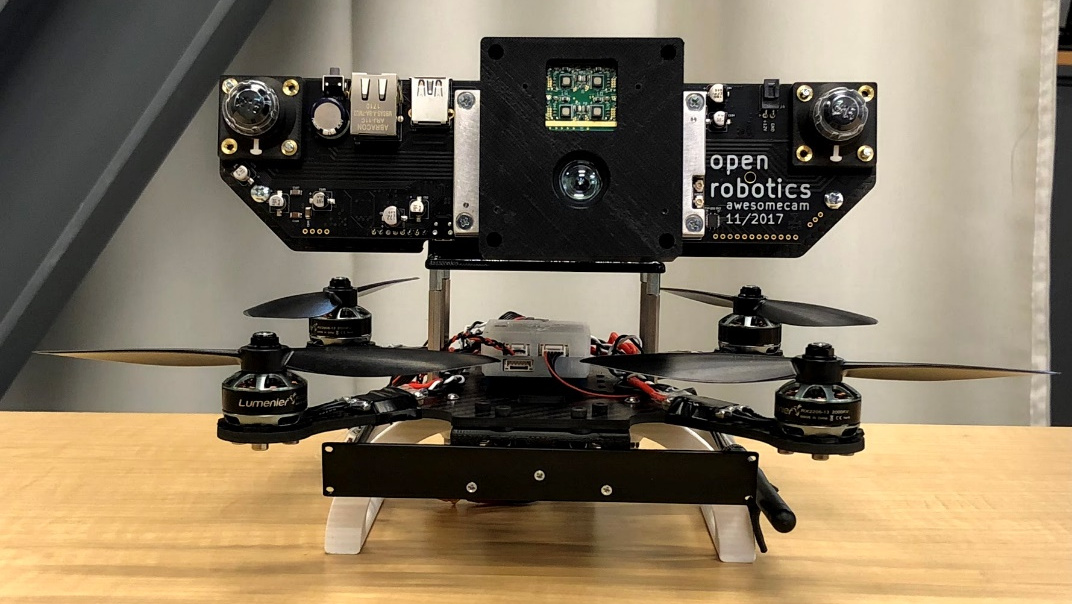}
\includegraphics[height=80px]{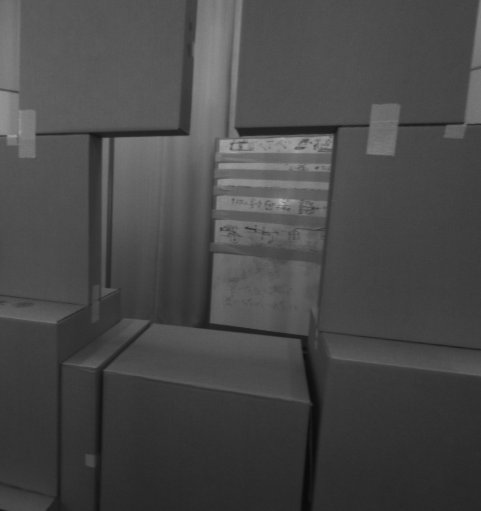}\\
\includegraphics[width=0.45\linewidth]{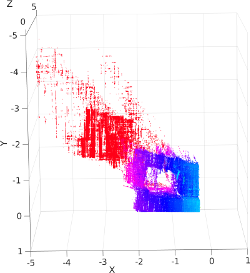}
\includegraphics[width=0.45\linewidth]{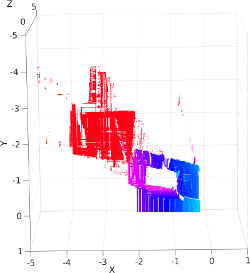}
\centering 
\caption{a) Top-L: Our stereo camera with a PMD Monstar placed roughly in the middle for optimal overlap with the stereo imagers, mounted on our Falcon 250 UAV platform. b) Top-R: A grayscale image collected from our stereo setup c) Bottom-L: Corresponding point cloud generated using Semi Global Matching d) Bottom-R: Corresponding point cloud generated using our Diffusion based approach - by fusing the two sensors, we can obtain high resolution depth estimates that are robust to the noisy measurements that is often seen in regular stereo based depth estimation. Colors are mapped between 0 - 2.5 meters. The red surface represents the curtain viewed through the aperture in the cartons (in blue)}
\label{fig:robot_and_pc}
\vspace{-0.75cm}
\end{figure}

LIDAR sensors are a popular choice for accurate and efficient depth estimation. These sensors are expensive and are often heavy to mount on a small robot platforms such as an unmanned aerial vehicle. Time-of-flight devices such as the PMDTec Monstar \cite{kraft20043d} provide dense depth estimates at a low resolution. They provide accurate short range depth measurements and are often used in indoor robotics and low light environments. Unlike photogrammetry based stereo, this sensor performs very well on surfaces with uniform appearance such as flat plain walls. Phase difference between the emitted and returned infrared signals are used to measure distances, and sensors such as the PMD are a practical alternative that we have successfully used on our unmanned aerial vehicle platform for indoor obstacle avoidance and mapping. (Fig \ref{fig:robot_and_pc})

\begin{figure*}[t!]
\centering
\includegraphics[width=81px]{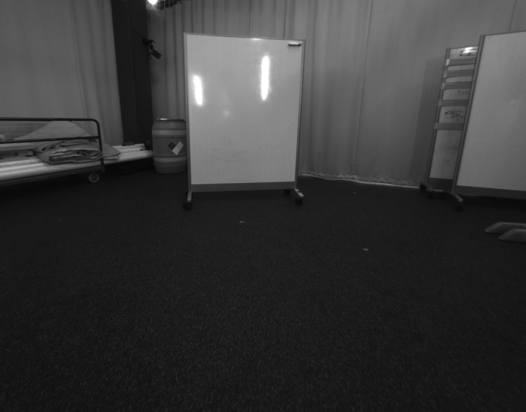}
\includegraphics[width=81px]{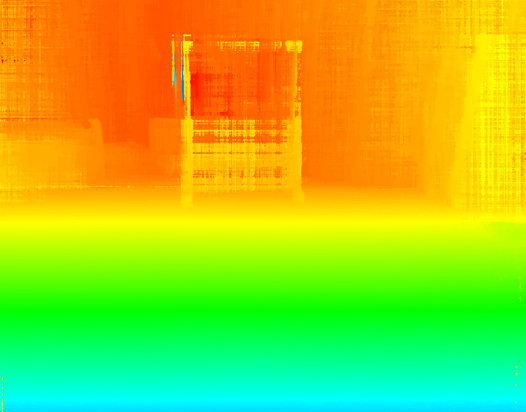}
\includegraphics[width=81px]{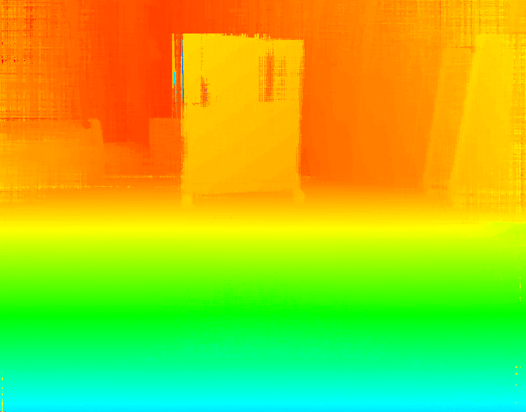}
\includegraphics[width=81px]{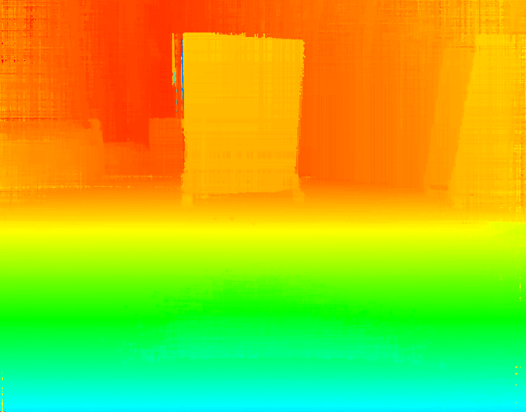}
\includegraphics[width=81px]{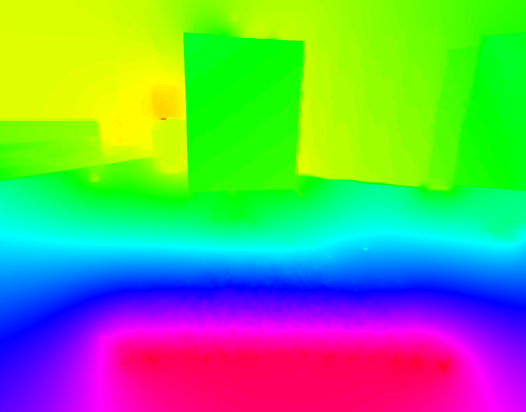}
\includegraphics[width=81px]{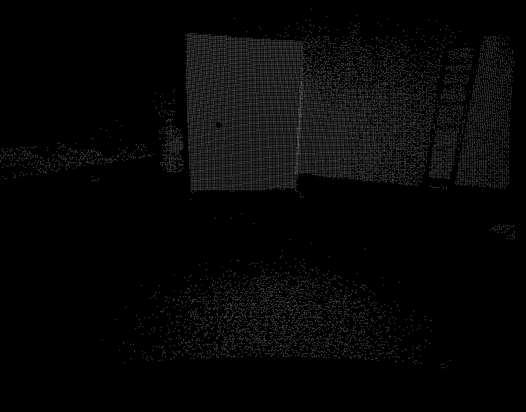}\\
\vspace{0.05cm}
\includegraphics[width=81px]{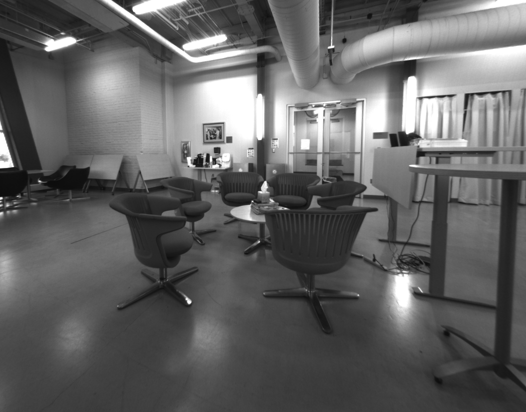}
\includegraphics[width=81px]{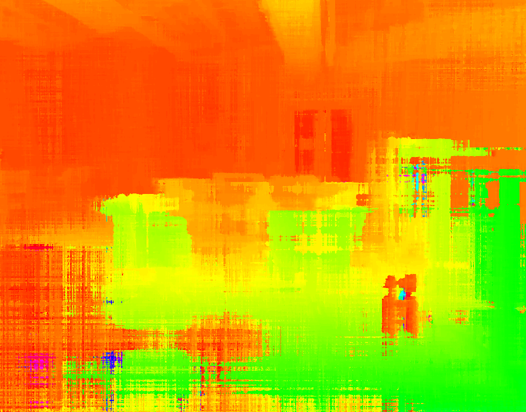}
\includegraphics[width=81px]{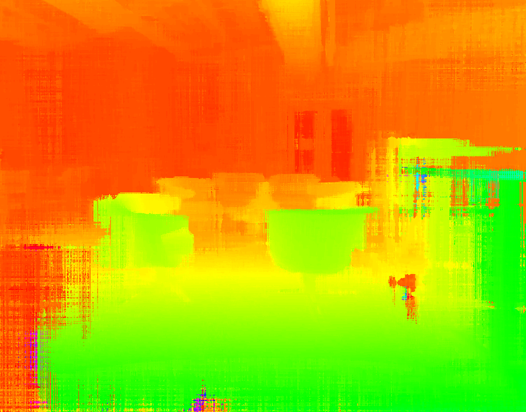}
\includegraphics[width=81px]{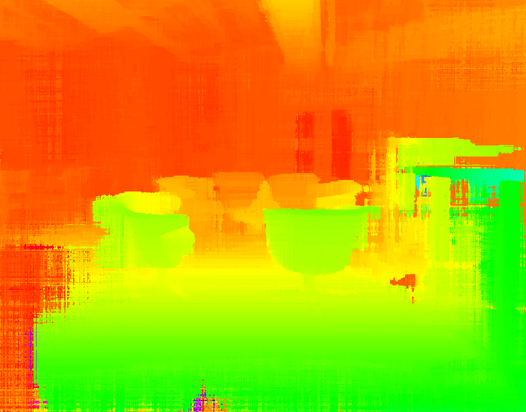}
\includegraphics[width=81px]{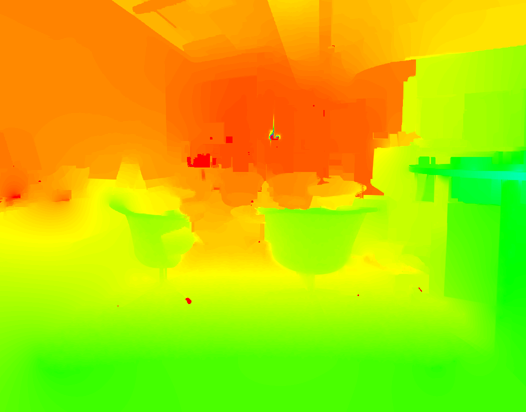}
\includegraphics[width=81px]{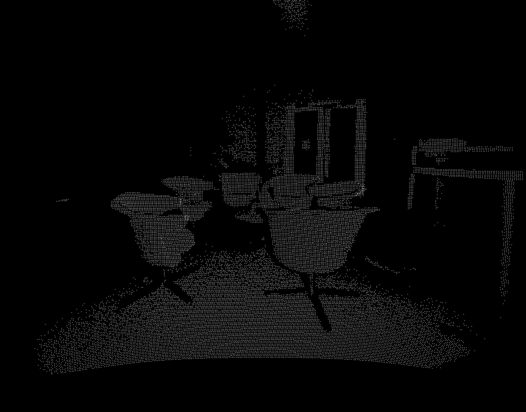}\\
\vspace{0.05cm}
\includegraphics[width=81px]{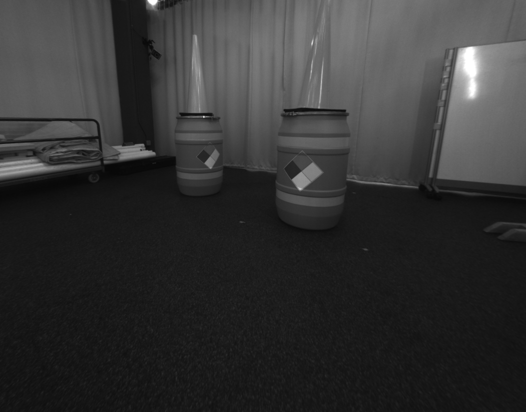}
\includegraphics[width=81px]{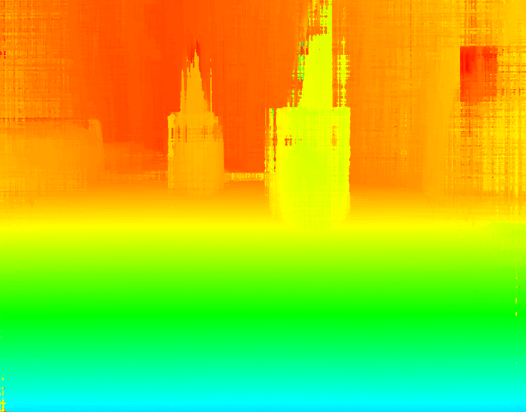}
\includegraphics[width=81px]{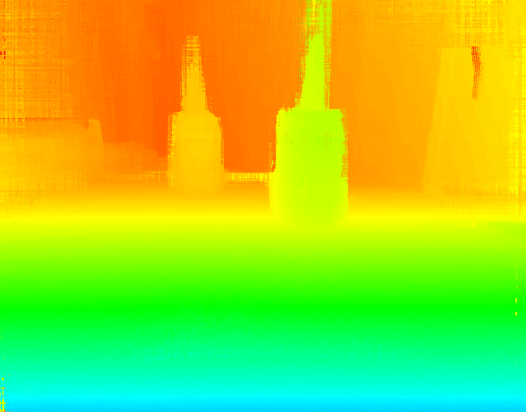}
\includegraphics[width=81px]{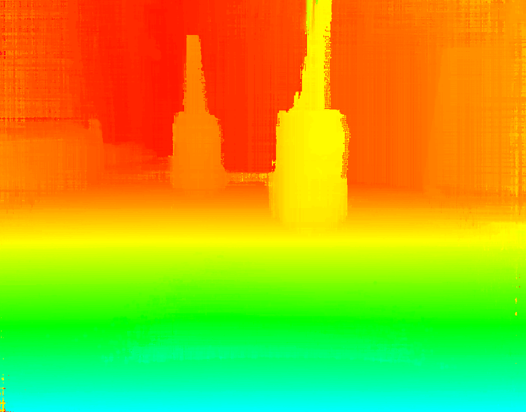}
\includegraphics[width=81px]{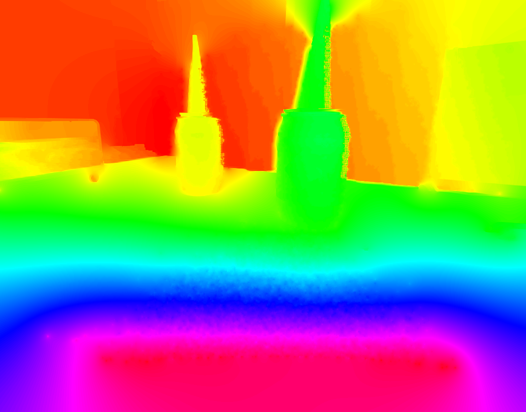}
\includegraphics[width=81px]{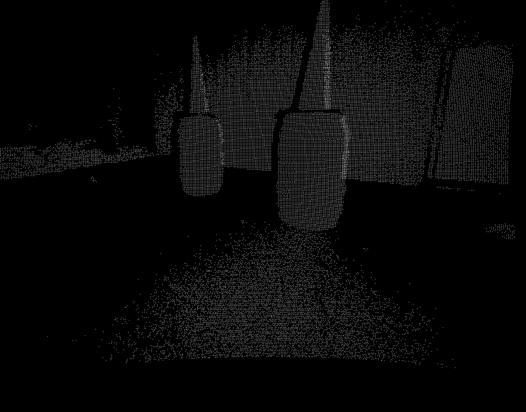}\\
\vspace{0.05cm}
\includegraphics[width=81px]{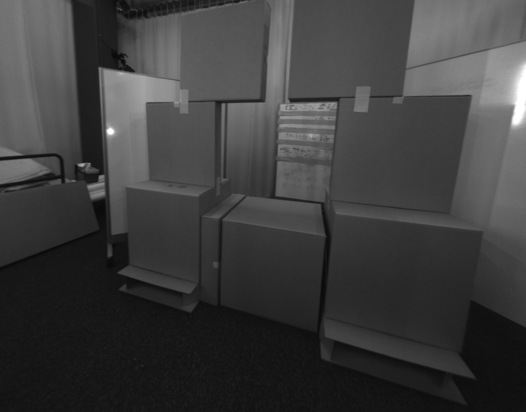}
\includegraphics[width=81px]{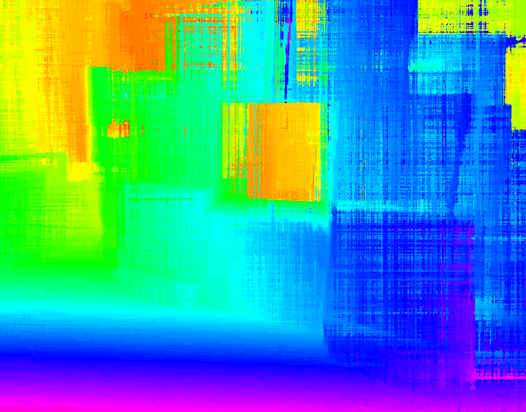}
\includegraphics[width=81px]{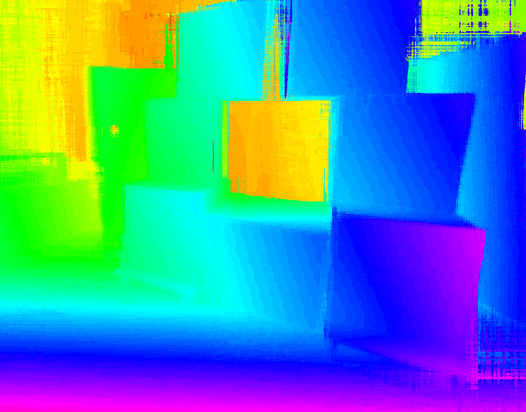}
\includegraphics[width=81px]{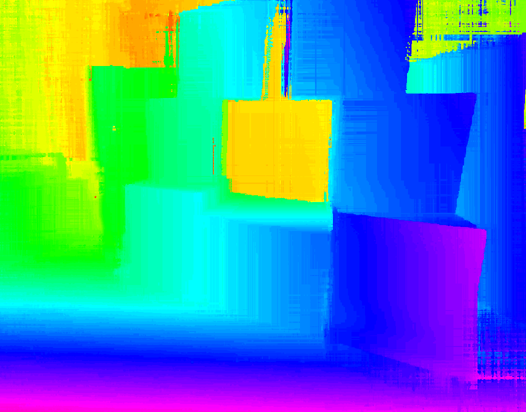}
\includegraphics[width=81px]{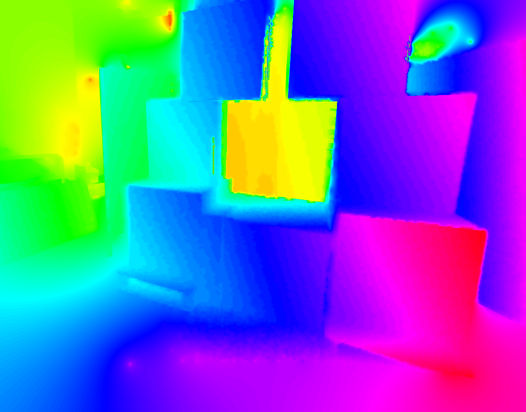}
\includegraphics[width=81px]{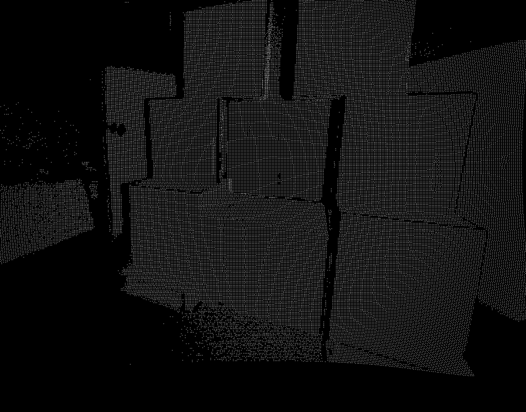}\\
\caption{(PMD Monstar Dataset) L-R: Rectified left image (grayscale); Results of the Semi-Global Matching - the algorithm performs poorly on uniform textures such as the whiteboard and the cardboard cartons, common to most stereo algorithms; Neighborhood Support method - performs slightly better than SGM but still has trouble filling in erroneous stereo estimates; Diffusion based method - performs significantly better than both the above methods, while preserving disparity discontinuities and retaining accurate disparity measurements; Anisotropic Diffusion - performs well at regions where PMD measurements exist, but greatly misrepresents disparity measurements at regions far away from such points, as one could expect from a monocular only setup; PMD Monstar points used after filtering out noisy measurements}
\label{fig:monstar_output}
\vspace{-0.5cm}
\end{figure*}

Beder et al. compare both sensing approaches under optimal conditions and concluded that the PMD performed with better accuracy in surface reconstruction, but that an ideal setup would be a fusion of both systems to overcome the low resolution of the PMD \cite{beder2007comparison}. Scharstein et al. discuss several different approaches to stereo depth estimation \cite{scharstein2002taxonomy}; though slightly dated, this work still presents a comprehensive list. Modern approaches use Convolutional Neural Networks to solve the disparity estimation and patch matching problem, however these networks are often too computationally demanding to run in real-time at our desired resolution \cite{zbontar2016stereo}. We build our method upon the seminal work of Hirschmuller et al. on Semi Global Matching (SGM) stereo algorithm, a method widely popular even today. \cite{hirschmuller2005accurate}.

Huber et al. present a lidar and stereo integration \cite{huber2011integrating} method that reduces the disparity search space and reduces computation time by a factor of five but assumes the lidar measurements are almost uniformly distributed over the image. Their second method, a dynamic programming based approach, uses the lidar points in the optimization process, however no quantitative results are provided for comparison with ours. Maddern et al. propose a probabilistic method of fusing sparse 3D lidar data with dense stereo data by modeling the error characteristics of both sensors and deploy it on a low-power GPU system \cite{maddern2016real}. They use lidar measurements as support points for the stereo method mentioned in \cite{geiger2010efficient}.
Veitch-Michaelis et al. also propose a similar approach with a region growing based stereo algorithm \cite{veitch2015data}.

A survey of ToF-stereo fusion is presented by Nair et al. \cite{nair2013survey} and we describe some of these solutions. Kuhnert et al. introduce an early variant of the PMD sensor and present a direct approach of fusing the PMD depth estimates with a WTA-style stereo algorithm \cite{kuhnert2006fusion}. Kahne et al. present a PMD and dense stereo fusion pipeline, where PMD measurements are used to reduce the disparity search space and the disparity estimation is set up as a graph cuts problem \cite{hahne2008combining}. This work does not present any quantitative estimates of accuracy improvements or computational feasibility for real time systems. Zhu et al. present the ToF-stereo fusion problem as a Belief Propagation problem using Markov Random Fields, where the weights correspond to confidence in each sensor's measurements \cite{zhu2008fusion}. The work of Gandhi et al. \cite{gandhi2012high} presents a similar ToF-stereo fusion pipeline, where the ToF depth points are seeds in a stereo seed growing algorithm. Their pipeline is validated on their own dataset as well as the Middlebury dataset, where points are uniformly sampled from the ground truth.
Gudmundsson et al. use a ToF camera and use the range values to constrain a hierarchical stereo matching algorithm \cite{gudmundsson2008fusion}. 

A different, but relevant method involves using the range measurements along with a monocular image to generate dense depth by guided interpolation. Courtois et al. use bilateral filtering based interpolation of lidar data for the purpose of robot mapping \cite{courtois2017fusion}. This paper presents quantitative results on the KITTI dataset, and compares their method to previous methods. Prembida et al. propose an interpolation and up-sampling based method of obtaining dense disparity estimates from LIDAR scans \cite{premebida2016high}. Ma et al. propose a deep learning approach to depth up-sampling and are currently in the top 3 on the KITTI depth completion benchmark \cite{ma2017sparse} \cite{uhrig2017sparsity}.

Fischer et al. present work that is most similar to ours, where ToF depth information is filtered and used to guide the cost aggregation stage of the Semi-Global Matching algorithm \cite{fischer2011combination}. The proposed method uses ToF depth to limit the search space during pixel-wise matching if the ToF data is in approximate agreement of the na\"ive matching estimate. This approach is intuitively similar to our neighborhood support mentioned in Section II. 

The main \textbf{contributions} of our work are: \textit{(1)} a method of integrating sparse accurate depth measurements into a dense stereo depth estimation framework, combining traditional stereo range fusion and depth interpolation techniques; \textit{(2)} a quantitative evaluation of our method on KITTI 2015 and Middlebury 2014 datasets and a small computational footprint allowing real time dense depth estimation on computationally constrained mobile robots.

\section{TECHNICAL APPROACH}

\subsection{Pipeline}
Our processing pipeline obtains as input a pair of rectified stereo images $I$ (left camera) and $J$ (right camera). Additionally, using the calibrated intrinsics and extrinsics, we convert the depth sensor's range measurements into a depth image in the left camera's reference frame with  matching focal length.

\textit{Setting up the cost volume}: As is common in many stereo algorithms, we first transform our grayscale intensity image to a feature space more robust to intensity variations. We apply the census transform to a window around each pixel in the left and right image and the resulting bitvectors are denoted by $I_{\textnormal{cen}}(x,y)$ and $J_{\textnormal{cen}}(x,y)$ \cite{zabih1994non}. A 3D cost volume is then computed, where the $X$ and $Y$ axes correspond to the 2D image co-ordinates and the $Z$ axis to the disparity range. Each element of this volume $C((x,y),d)$ represents a cost, or similarity between the transformed value in the left image and its corresponding value in the right, displaced in the y-axis by $d$, where $d=1..D_{MAX}$ as seen in Eq \ref{eq:cost_volume},
%
%
\begin{equation}
C((x,y),d) = SIM(I_{\textnormal{cen}}(x,y),J_{\textnormal{cen}}(x,y-d))
\label{eq:cost_volume}
\end{equation}
Here, the similarity measure $SIM(a,b)$ is the Hamming distance between the two census bit vectors from the left and right images.

\textit{Cost aggregation}: We follow the aggregation method proposed by Hirschmuller et al. \cite{hirschmuller2005accurate}. The aggregation step is formulated as an energy minimization equation, reminiscent of scanline optimization based stereo methods, done along multiple different directions at every pixel. To reduce the computational complexity, we consider 4-8 directions, instead of the original 16. The SGM algorithm proceeds by computing aggregate costs along a number of different directions as described in Eq \ref{eq:sgm_agg}.
\begin{equation}
C_{r}'(p,d) = C(p,d) + min
\begin{cases} 
C_{r}'(p-1,d)\\
C_{r}'(p-1,d+1) + P_{1}\\
C_{r}'(p-1,d-1) + P_{1}\\
min_{i}C_{r}'(p-1,i) + P_{2} 
\end{cases}
\label{eq:sgm_agg}
\end{equation}

Where $C_{r}'$ is the aggregated cost volume for a given path $r$ and $p$ indicates a point along a path $r$. The notation $p-1$ indicates the point previous to $p$ along the direction $r$. The penalty terms $P_{1}$ and $P_{2}$ indicate how heavily to penalize small disparity differences and large ones respectively. The final disparities are calculated by summing over the different paths $r$ and selecting the disparity level $d$ with the lowest cost.
\begin{equation}
D(x,y) = \operatorname{arg\,min}_dS((x,y),d)
\end{equation}
\begin{equation}
S((x,y),d) = \sum_{r}C_{r}'((x,y),d)
\end{equation}
We keep this energy minimization formulation as is, and seek to introduce the depth measurements during the cost volume construction phase.

\textit{Updating the cost volume:} We introduce our range measurements at the cost volume creation stage, making updates to element $((x,y),z)$ in the cost volume at points where there is a measured disparity value. We denote these elements by $((x_{m},y_{m}),d_{m})$. These measured readings are treated as high confidence disparity estimates and are used to modify the original cost volume entries. We propose three different approaches,

\subsubsection{Na\"ive Fusion}
The na\"ive approach involves setting the cost at the measured point $(x_{m},y_{m},d_{m})$ to be a very small value, 0. Intuitively, for a given pixel this means that we have the highest confidence that the disparity to be assigned to this point is $d_{m}$.
\begin{equation}
C((x_{m},y_{m}),d_{i}) =
\begin{cases} 
0 \textrm{ if $d_{i}=d_{m}$}\\
C((x_{m},y_{m}),d_{i})\textrm{ otherwise}\\
\end{cases}
\end{equation}

By na\"ively altering the cost elements $((x_{m},y_{m}),d_{m})$, we see an improvement over the basic SGM algorithm, however, the nature of the cost aggregation makes it robust to points that are in strong disagreement with their neighbors, both spatially and along the disparity axis. Therefore intuitively, the aggregation procedure tries to reject or ignore very sparse updates made to the cost volume that are in strong disagreement with the original stereo costs. Additionally, the spread of information is limited to the paths along which an update was made, and if the range measurements are too sparse, significant improvement is not observed.

\subsubsection{Neighborhood Promotion}
As a solution to the above problem, we propose the following method: Since we are confident about the range measurements that we obtain from our range sensor, we can force lowered costs or energies on points in the image that neighbor these sparse locations, essentially providing more guidance for the energy minimization. We use the grayscale image as the guide, assuming that within small windowed regions, the grayscale intensities of two points on a surface having similar depth also have similar intensities. For every range measurement $((x_{m},y_{m}),d_{m})$, we observe the grayscale intensities of its neighbors and calculate a set of weights based on the intensity difference between the point $(x_{m},y_{m})$ and its neighbors, within a window of radius $K_{w}$. The weight matrix $W_{m}(i,j)$ is calculated using a Gaussian with a smoothing parameter $\sigma_{r}$.
\begin{equation}
W_{m}(i,j) = G_{\sigma_{r}}(I(i,j)-I(x_{m},y_{m}))
\end{equation}
Therefore for each window region, a cost update is made to all pixels in this region. For each disparity level $k$, the cost updates are as follows,
\begin{equation}
C((x_{m},y_{m}),d_{k}) =
\begin{cases} 
\beta \textrm{ if } |d_{k}-d_{m}| \geq \tau_{d}\\
\epsilon \textrm{ otherwise}\\
\end{cases}
\end{equation}

And the cost update to the neighboring pixels is,
\begin{equation}
C((x_{i},y_{j}),d_{m}) =
\begin{cases} 
(1-W_{m}(i,j))\beta \textrm{ if } W_{m}(i,j) < \tau_{n}\\
\epsilon \textrm{ otherwise}\\
\end{cases}
\end{equation}

where $(i,j)$ are co-ordinates for the points with respect to $(x_{m},y_{m})$ within the window region of radius $K_{w}$. Notice that we now also introduce a parameter $\tau_{d}$, which controls the degree of belief in our measured disparity accuracy. And to further propagate our belief in our measured disparity $d_{m}$, we set all other disparity costs to be some large constant $\beta$. Here $\epsilon$ is the minimum cost assigned. The cost update made along the Z-axis between $k$ and $\tau_{d}$ can be modeled either by some prior noise model associated with the sensor or by assigning constant values or linear gradients. We observed that a good value for $\tau_{d}$ is 2 for the PMD measurements and that a smooth interpolation of costs between $\tau_{d}$ and $k$ does not show substantial improvement versus a constant cost update for small values of $\tau_{d}$. Similarly, the threshold $\tau_{n}$ determines how similar two intensity values must be in order to believe that they are part of the same surface.

\subsubsection{Diffusion Based Update}
%
%
%
With the intuition that larger support regions provide more information to the optimization procedure, we propose a third approach based on Anisotropic Diffusion and depth interpolation to update the volume \cite{perona1990scale}. Intuitively, we interpolate the sparse depth points from the range sensor and then use this information during the update step. We restrict this interpolation to regions near valid range measurements as points further away may not be part of the same surface. When using a PMD sensor, the points measured are usually close to the robot and trying to interpolate values far away leads to large inaccuracies in the resulting disparity.

Our interpolation limits are defined by a radius $K_{\textnormal{interp}}$ around each measured point. Points with valid measurements remain unaltered and the remaining points are assigned disparity values by leveraging the grayscale images for interpolation. Each interpolated disparity value is a weighted combination of all of the sparse disparity measurements within a radius of $K_{\textnormal{interp}}$ pixels of that location. 

\begin{equation}
D(x,y) = \frac{\sum_{i,j}W(i,j)d(i,j)}{\sum_{i,j}W(i,j)}
\end{equation}
where $(i,j)$ are co-ordinates of measured disparity estimates from the sensor that are within $K_{\textnormal{interp}}$ pixels from $(x,y)$. The weights $W(i,j)$ are calculated using a bilateral filtering method while $d(i,j)$ are disparity measurements from the range sensor \cite{tomasi1998bilateral}. Bilateral filters are commonly used edge preserving filters and in our case, depth discontinuity preserving. The pixel distance based smoothing  is parameterized by $\sigma_{d}$.
\begin{equation}
\begin{split}
W(i,j) &= G_{\sigma_{r}}(I(i,j)-I(x_{m},y_{m})) \\
	   &\times G_{\sigma_{d}}(|(i,j)-(x_{m},y_{m})|)
\end{split}
\end{equation}
For computational efficiency, we compute $\sum_{i,j}W(i,j)d(i,j)$ and $\sum_{i,j}W(i,j)$ separately and perform the division upon completion. Additionally, this can be computed in parallel over the set of points within the $K_{\textnormal{interp}}$ region around each measured point. Maintaining the sum of the weights also provides us with additional information regarding how similar a point at $(x,y)$ is to its nearest measured point $(x_{m},y_{m})$ which serves as a proxy for our confidence in the interpolated disparity value. The update step is then,
\begin{equation}
C((x_{i},y_{j}),d_{k}) =
\begin{cases} 
(1-W(i,j))\gamma \quad \textrm{if } \tau_{l} < W(i,j) < \tau_{u}\\
\gamma \quad \textrm{if }W(i,j) \leq \tau_{l}\\
\epsilon \quad \textrm{if }W(i,j) \geq \tau_{u}\\
\beta \quad \textrm{if } |d_{k}-d_{v}| \geq \tau_{d} \textrm{ and } W(i,j) > \tau_{l}\\
\end{cases}
\end{equation}

Here, we use $d_{v}=D(x_{i},y_{j})$ to refer to interpolated depth sensor disparities at the point $(x_{i},y_{j})$. The parameter $\tau_{m}$ controls our confidence in the interpolated value, using weight $W(i,j)$. Parameters $\epsilon$, $\beta$ and $\tau_{d}$ remain the same as before. Here, $\gamma$ is some large penalty, similar or equal to $\beta$. The parameters $\tau_{u}$ and $\tau_{l}$ indicate a confidence range over the normalized weights $W$, representing a cutoff for high confidence and low confidence respectively.

\section{RESULTS AND ANALYSIS}

\begin{figure*}
\centering
\includegraphics[width=96px]{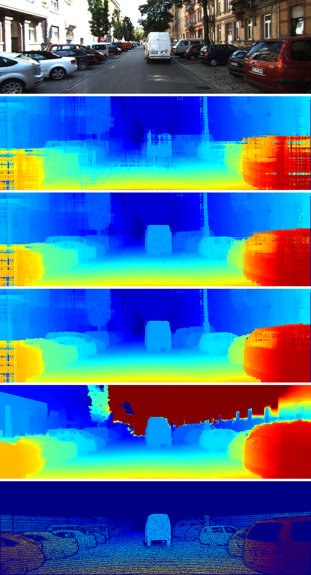}
\includegraphics[width=96px]{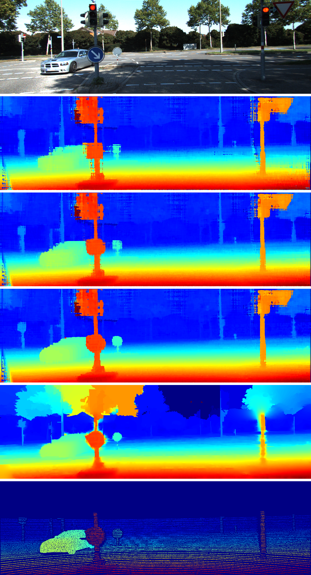}
\includegraphics[width=96px]{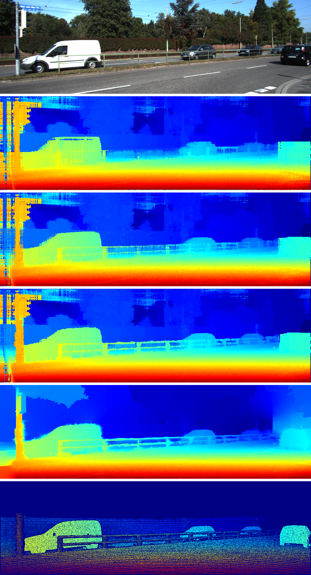}
\includegraphics[width=96px]{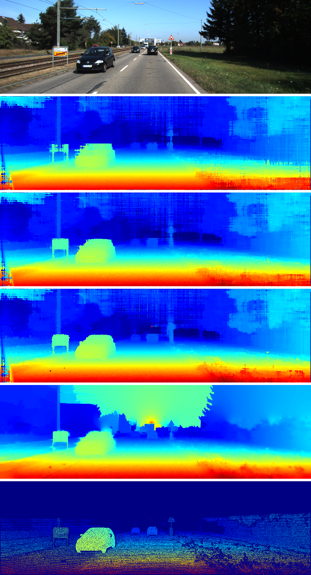}
\includegraphics[width=96px]{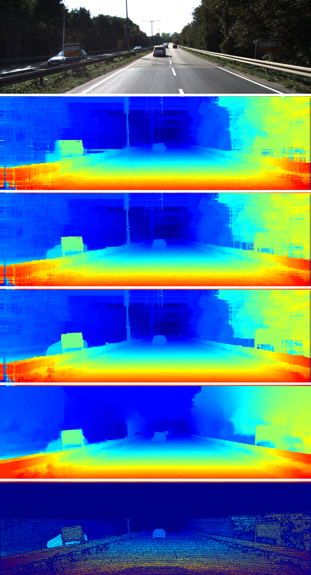}
\caption{(KITTI 2015) T-B: Rectified color image; Results of Semi Global Matching; Neighborhood Support method; Diffusion based method; Anisotropic Diffusion - note that the algorithm struggles to extrapolate a disparity value at regions where no seed range points exist and fails completely in certain cases; Ground truth points - these are the points at which the algorithm is evaluated, this is the original ground truth data with our sampled seed points removed. For these illustrations that is 15\% of the total points available.}
\vspace{-0.5cm}
\label{fig:kittiimages}
\end{figure*}

We discuss the performance of our algorithm on three datasets providing quantitative results on the KITTI 2015 and Middlebury 2014 datasets and qualitative results on our own PMD Dataset. On each dataset, we compare to the standard \textit{Semi Global Matching} algorithm, without the left-right consistency check. For consistency we use the same $P_{1}$ and $P_{2}$ parameters for all images within a dataset, for all methods relying on SGM. Across datasets, we manually select $P_{1}$ and $P_{2}$ values after a parameter search. We choose $D_{MAX}$ to be $256$, which is sufficient for both our dataset as well as KITTI. On Middlebury 2014 there are some points with larger disparity values but we only evaluate points within this range. 

We also compare against an \textit{Anisotropic Diffusion} based approach to depth enhancement \cite{liu2013guided}. It must be noted that this method is independent of stereoscopic information and takes as its input a single image and a set of disparity points and generates a dense disparity image by diffusing these disparity points, using the input image as a guide. 

\begin{table}
\begin{center}
\caption{KITTI 2015 results}
\label{table:kitti_errors_T}
Each element in the table represents the percentage of pixels with disparity error greater than $1$, $2$ and $3$ disparities away from ground truth.
\vspace{0.2cm}

 \begin{tabular}{|c| c c c|} 
 \hline
 \textbf{Method} & \textbf{$>$1px} & \textbf{$>$2px} & \textbf{$>$3px} \\ 
 \hline
 SGM & 17.10 & 10.60 & 7.73 \\ 
 \hline
 Na\"ive & 15.77 & 9.88 & 7.22 \\
 \hline
 Neighborhood Support & 12.93 & 4.09 & 2.59 \\
 \hline
 Diffusion Based & \textbf{4.26} & \textbf{2.01} & \textbf{1.51} \\
 \hline
 Anisotropic Diffusion & 5.56 & 3.52 & 2.58 \\ 
 \hline
\end{tabular}
\end{center}
\vspace{-0.5cm}
\end{table}

\begin{figure}
\centering
\includegraphics[width=\linewidth]{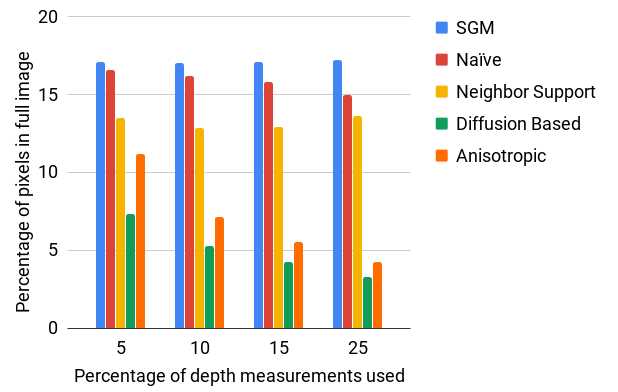}
\caption{Plot of error rates ($>$1px) versus number of samples for 5, 10, 15 and 25\% of available ground truth lidar measurements. Evaluation is performed on the remaining samples.}
\vspace{-0.6cm}
\label{graph:samples}
\end{figure}

\subsubsection{KITTI 2015 Dataset}
We evaluate these methods on the 200 stereo pairs provided. However, the ground truth data is not for every pixel in the grayscale image, but an accumulation of Lidar points over a range of frames before and after the reference image. For our evaluation, we use a subset of these measurements as our sparse depth input along with the stereo pair and our evaluation is done on ground truth measurements outside of this sample set. We randomly sample 15\% of the ground truth depths for our final evaluation. We use the development kit provided with the dataset and report our errors from one to three disparity values, with a small tolerance as specified. The error rates are shown in Table \ref{table:kitti_errors_T}. A few example disparity maps are shown in Figure \ref{fig:kittiimages}. Therefore, it is clear that our \textit{Diffusion Based} method outperforms the others. Interestingly, \textit{Anisotropic Diffusion} performs reasonably well. But this can be attributed to the fact that evaluation is performed on points where ground truth exists, and this is spatially in proximity to where our points are sampled from, even though randomly sampled. This algorithm however suffers from the interpolation flipping problem previously mentioned, as can be seen in Fig \ref{fig:kittiimages} a,b,d. The na\"ive approach does not significantly outperform \textit{SGM}, and thus reaffirms our observation that creating larger areas of low energy around measured disparity levels in the cost volume is important for improving performance. 

We also plot the error rates, which increase in the number of sampled points. This is shown in Figure \ref{graph:samples}. An interesting observation is the increase in error at 25\% samples for the \textit{Neighborhood Support} method. This is because for a fixed window size, an increase in the density of samples, will cause conflicting update to neighborhood regions, a problem that we solve in the \textit{Diffusion Based} method by taking a weighted average of all support measurements. At the time of writing, the current state of the art on the KITTI 2015 Stereo benchmark achieves an error of 1.74\% in the 3px error range. Though not directly comparable, we are able to achieve similar performance with our \textit{Diffusion Based} method, scoring 1.51\% on the training dataset provided, using only a small fraction of the lidar measurements.

\begin{figure*}[t!]
\centering
\includegraphics[width=\linewidth]{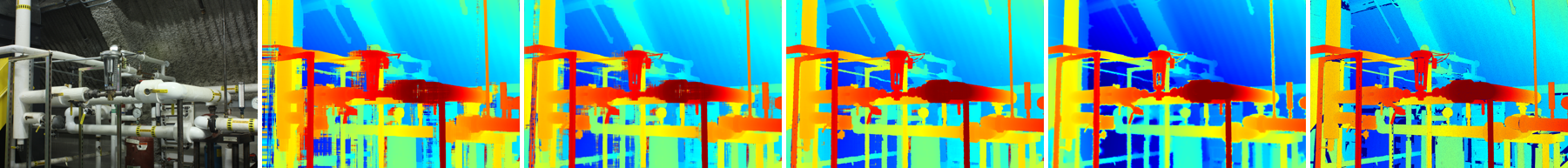}\\
\vspace{0.05cm}
\includegraphics[width=\linewidth]{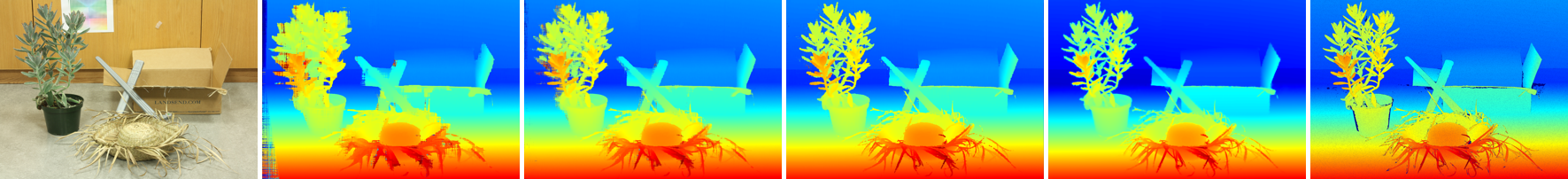}\\
\vspace{0.05cm}
\includegraphics[width=\linewidth]{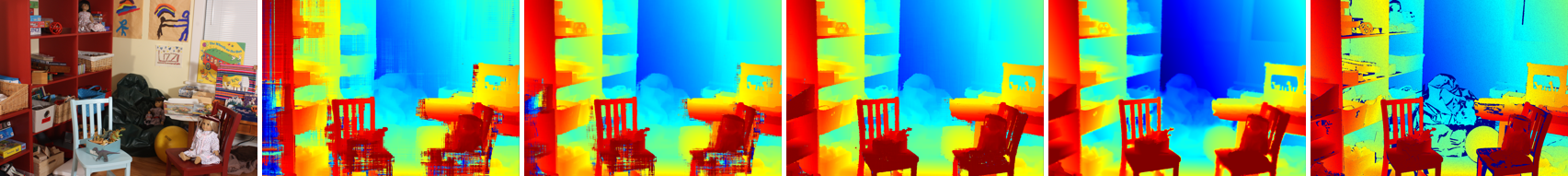}\\
\vspace{0.05cm}
\includegraphics[width=\linewidth]{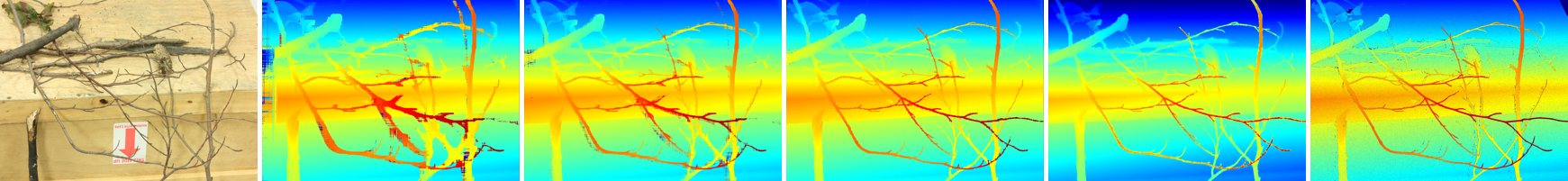}\\
\vspace{0.05cm}
\includegraphics[width=\linewidth]{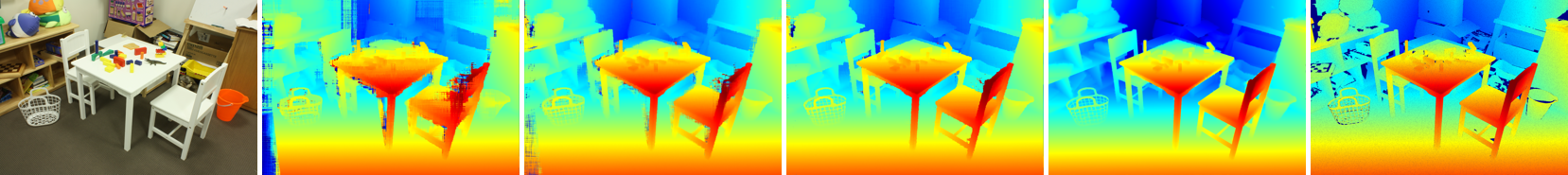}\\
\vspace{-0.1cm}
\caption{(Middlebury 2014) L-R: Left rectified \textit{image} (full resolution), \textit{Semi-Global Matching} results - same $P_{1}$ and $P_{2}$ values were used for all images in this dataset and for all following methods. Here $D_{MAX}$ was chosen to be 256; \textit{Neighborhood Support} method - shows better performance when compared to SGM and our na\"ive approach (not pictured), but this algorithm still has problems with edges and partial occlusions; \textit{Diffusion based} - this method performs the best, showing robustness to noise and preserving edges even with a small number of points; \textit{Anisotropic Diffusion} - this method performs well at preserving edges and interpolating disparities, but regions where no nearby measurements exist result in incorrect disparities; \textit{Ground Truth}.}
\label{fig:middlebury_images}
\vspace{-0.5cm}
\end{figure*}

\subsubsection{Middlebury 2014 Dataset} 
For Middlebury 2014, we evaluate our algorithm on 23 of the stereo pairs provided, which have dense ground truth measurements. Since this dataset provides accurate, dense ground truth, we sample 2.5\% of the total ground truth points, and randomly add noise to the measurements (up to 5\%). These measurements are then used as before. As can be seen in Table \ref{table:errors_mb} the \textit{Diffusion Based} method achieves the lowest error rate, which can also be qualitatively seen in Fig \ref{fig:middlebury_images}. The anisotropic diffusion method also yields impressive results, but sees significant error due to misinterpreting depth boundaries on appearance alone. The \textit{na\"ive} method doesn't show significant improvement over \textit{SGM}, while the neighborhood support method performs notably better. 
\begin{table}
\begin{center}
\caption{Middlebury 2014 results}
\label{table:errors_mb}
Each element in the table represents the percentage of pixels with disparity error greater than $1$ disparity away from ground truth.\\
\vspace{0.2cm}
 \begin{tabular}{|c| c|} 
 \hline
 \textbf{Method} & \textbf{$>$1px} \\ 
 \hline
 SGM & 3.4037 \\ 
 \hline
 Na\"ive & 3.2801 \\
 \hline
 Neighborhood Support & 1.8067 \\
 \hline
 Diffusion Based &  \textbf{0.1921}\\
 \hline
 Anisotropic Diffusion & 0.4297 \\ 
 \hline
\end{tabular}
\end{center}
\vspace{-0.8cm}
\end{table}

\subsubsection{PMD Monstar Dataset}
For our dataset, we use a pair of Python1300 CMOS sensors with 2.8mm, 1/2in sensors and FOV 95.3 x 82.6 degrees. The PMD has resolution of 352 x 287, and FOV of 100x85 degrees and is shown in Fig \ref{fig:robot_and_pc}. The stereo camera pair is calibrated with the PMD to obtain accurate intrinsic and extrinsic estimates for all sensors. The stereo baseline is 20cm, with the PMD placed in the center. Range measurements from the PMD are projected onto the stereo cameras and depth measurements are transformed to disparity estimates in the stereo domain.

Since we lack ground truth information to verify our claims, we qualitatively discuss performance on this data. We notice a consistent improvement in and around regions where PMD measurements exists. Depth discontinuities are more accurate and edges are well preserved in both the Diffusion Based method as well as the Neighborhood Support method. We show these results in Fig \ref{fig:monstar_output}. Our \textit{Diffusion Based} method is able to effectively use PMD measurements from surfaces such as the white-board in Fig \ref{fig:monstar_output}a. Similar improvement is seen in the office setting image Fig \ref{fig:monstar_output}b, where \textit{SGM} struggles to assign correct disparities to the floor and chair regions. An extreme example is seen in Fig \ref{fig:monstar_output}d where \textit{SGM} incorrectly estimates disparities on the cartons and the white-board. \textit{Anisotropic Diffusion} performs well, having only monocular and range information to work with, but suffers in regions where nearby range information doesn't exist. Since it is heavily influenced by image intensity gradients, edges translate to depth discontinuities, and this becomes a problem when no range information exists nearby, causing the interpolated disparity to ambiguously flip, introducing depth discontinuities where none exist.

\section{CONCLUSIONS}


In summary, the present work explores several different means of incorporating sparse depth sensor measurements into a dense stereo algorithm. We evaluate different approaches on the KITTI and Middlebury 2014 datasets, and demonstrate how they improve upon an image-only based stereo vision approach. Na\"ively incorporating the depth data holds only marginal advantage, but propagating depth data points to neighboring regions yields much improved results. The lowest error statistics, as well as good qualitative results, are obtained by combining a census based stereo cost function with an image edge preserving interpolation of the depth measurements, followed by the SGM procedure. Using the same approach in a monocular setting is demonstrated to suffer from serious artifacts, highlighting the importance of utilizing the stereo disparities. Future work includes modeling of sensor and calibration noise characteristics to adaptively select confidence thresholds.

\balance 









\bibliographystyle{IEEEtran}
\bibliography{ref}

\end{document}